\documentclass[conference]{IEEEtran}
\usepackage{cite}
\usepackage{amsmath,amssymb,amsfonts}
\usepackage{algorithmic}
\usepackage{graphicx}
\usepackage{textcomp}
\usepackage{xcolor}
\usepackage{tabularx,booktabs}
\usepackage{multirow,tabularx}
\newcolumntype{C}{>{\centering\arraybackslash}X} % centered version of "X" type
\usepackage{lipsum}
\begin{document}

\title{Short-term prediction of photovoltaic power generation using Gaussian process regression}

\author{\IEEEauthorblockN{Yahya Al Lawati*, Jack Kelly+ and Dan Stowell*+}
\IEEEauthorblockA{*School of Electronic Engineering and Computer Science, 
Queen Mary University of London, London, UK \\
+Open Climate Fix, London, UK}
}

\maketitle

\begin{abstract}
Photovoltaic (PV) power is affected by weather conditions, making the power generated from the PV systems uncertain. Solving this problem would help improve the reliability and cost effectiveness of the grid, and could help reduce reliance on fossil fuel plants. The present paper focuses on evaluating predictions of the energy generated by PV systems in the United Kingdom Gaussian process regression (GPR). Gaussian process regression is a Bayesian non-parametric model that can provide predictions along with the uncertainty in the predicted value, which can be very useful in applications with a high degree of uncertainty. The model is evaluated for short-term forecasts of 48 hours against three main factors – training period, sky area coverage and kernel model selection – and for very short-term forecasts of four hours against sky area. We also compare very short-term forecasts in terms of cloud coverage within the prediction period and only initial cloud coverage as a predictor.
\newline
\end{abstract}

%\begin{IEEEkeywords}
%Gaussian process regression (GPR), photovoltaic (PV) power forecasting
%\end{IEEEkeywords}

\section{Introduction}
Fossil fuels are the primary source of energy worldwide, accounting for 84\% of primary energy use \cite{b1}. However, the world seeks alternatives, and renewable energy has gained interest: its consumption share has grown strongly to over 40\% (excluding hydroelectricity) of the global growth in primary energy in 2019 compared to 2018, with solar and wind power being the greatest beneficiaries \cite{b1}. The chief concern that accompanies solar energy is its uncertainty, which is mainly affected by weather conditions \cite{b2}. Being able to predict power generation mere hours ahead would help control the amount that must be generated from fossil fuels, reducing the amount of carbon dioxide (CO2) emissions produced. There are two main forecasting methods: model-based approaches and data-driven approaches. An example of model-based approaches is numerical weather prediction (NWP) which employs a set of equations to describe the flow of fluids. Model-based approaches can be complicated and computationally costly. Data-driven approaches, on the other hand, do not use any physical model. They are easier to implement and require no prior knowledge about weather forecasts. Forecasting techniques can be deterministic or probabilistic \cite{b2}\cite{b3}. Probabilistic approaches can predict a range of values with probabilities, and thus are more appropriate for forecasting applications.

In this work we evaluate the accuracy of short-term predictions of the amount of energy generated from photovoltaic (PV) systems using Gaussian process regression (GPR), trained using historical data of power output from PV systems together with cloud coverage obtained from high-resolution visible (HRV) satellite images.

%\section{Literature Review}
 
Integrating PV systems into the grid cannot be efficient unless there is reliable information about power forecast \cite{b7}. This is because, whilst it is possible to handle small energy drops by generating slightly energy more than demand, significant power drops must be replaced within seconds. Fossil fuel plants can take hours to spin from start, and storage battery devices are costly and inefficient for large quantities of photovoltaic panels \cite{b8,b9}. Short-term energy prediction can help to determine how much spinning reserve to schedule \cite{b8}. Spinning reserve is the unused power capacity which can be activated on the authority of the system operator to serve additional demand \cite{b10}. Numerical weather prediction models are useful for predicting elements such as temperature and wind speed, especially in the short term.  However, predicting solar irradiance can be inaccurate and very short-term predictions might be inapplicable, for several reasons \cite{b8,b11}:

\begin{itemize}
  \item Numerical weather prediction models require at least an hour to execute. This is a problem, especially for cloud forecasting, as it can change rapidly.
  \item There is no high demand for solar irradiance.
  \item Solar irradiance is computationally expensive to calculate.
\end{itemize}

The gap between the necessity and difficulty of predictions which use conventional forecasts opens the door for machine learning. Different solar forecasting models have been developed throughout the years, some of which have achieved great success in solar forecasting. Artificial neural networks \cite{b12,b13} and support vector machines \cite{b3,b14} are examples of popular models. Artificial Neural networks are known for their ability to model complex and non-linear relationships. For time-series applications Gaussian processes (GPs) appear to be recommended for their ability to handle complex relationships \cite{b10}. Our appendix explains GPs in more detail; we next discuss research conducted on GPs for PV modelling.

Weighted Gaussian Process Regression (WGPR) \cite{b2} assigns weights to each data sample, such that high outliers will have lower weights. The forecasting process involves multiple levels, and the model considers eight attributes: solar radiation, photosynthetic active radiation (PAR), ambient temperature, wind speed, gust speed, precipitation, wind direction, and humidity. The model has a training data period of approximately 33 days and a validation period of five consecutive days with five-minute intervals. Solar radiation and photosynthetic active radiation (PAR) were the most important predictors amongst the eight considered.

Gaussian process quantile regression (GPQR) model is discussed in \cite{b10,b14}. Quantile regression aims to estimate the quantiles of the conditional distribution of predicted values given predictor values, which makes the model more robust against outlier predictions. In addition, it captures the relationship between input and output variables. The GP model is trained with data from 20 days to predict the power load for the following seven days.
The results show that GPQR performed better than GPR by 0.26\% on average (using mean absolute percentage error, MAPE) for one-hour predictions and 0.49\% on average for two-hour predictions.

Other papers have discussed models based on GPs \cite{b16,b17,b18}.
Both the models we discussed show better results when compared to GPR in their own comparisons. However, the accuracy difference is not large, and the added sophistication may imply heavier computational demands.

According to \cite{b14}, clouds are one of the factors that have a significant influence on solar radiation.
It is essential to note that, other than the issues mentioned regarding NWP, cloud coverage data are parameterised, whilst depending on satellite images can provide more accurate readings \cite{b8}. Parameterisation means that the information is converted to an abstract value that presents general knowledge. For example, instead of resolving clouds, general information is provided, such as whether the day was sunny or cloudy.

The question raised in this paper is: ``Can a simple Gaussian Process Regression model accuracy enhance by depending on satellite images instead of Numerical Weather Predictions?'' This paper's contributions can be summarised as:
\begin{itemize}
  \item Exploring different factors that can enhance short-term forecasts (48 hours) according to satellite images.
  \item Exploring different factors that can enhance very short-term forecasts (four hours) according to satellite images at each point of the forecast and initial satellite images at the moment of the forecast.
  \item Comparing forecasts depending on cloud coverage at each point and the initial cloud coverage for very short-term forecasts (four hours). 
\end{itemize}

\section{Methodology}
The model was trained based on historical data and cloud coverage. The experiments were divided into two sets. The first set of experiments examined three factors in order to obtain the most accurate solar power forecasts for the following 48 hours: training period, sky coverage area and kernel structure. The second set of experiments examined very short-term predictions for the subsequent four hours in terms of sky coverage area. In addition, a comparison was made between forecasts with consideration of cloud coverage within the prediction period and forecasts with only initial cloud coverage provided. In addition, a third set of experiments was conducted to test the influence of the periodic kernel on model accuracy. However, the model failed to follow the trend of the actual power generated. Hence, this set of experiments is not discussed in the paper.\\
The results of the first and second sets evaluated amongst four different PV systems in the United Kingdom.
The paper assumes that the cloud forecast data will be provided either by cloud forecasting models or by prediction algorithms in case applied as a real-world application. This section describes two aspects: data preparation and model accuracy evaluation.

\subsection{Data Preparation}
The model uses historical data from given PV systems and cloud coverage to predict the amount of energy in the following four and 48 hours. It uses three data sources, including historical data on power generated from different systems and the metadata of those systems, such as location, system capacity and other system details. The third source is satellite images in five-minute intervals. The first step was to identify the geospatial boundary in transverse Mercator projection (metres) of the United Kingdom. The next step in the data preparation was to load PV system metadata and convert latitude and longitude to transverse Mercator projection. Then, systems outside the United Kingdom were identified and removed from the data. After that, data on energy generated from systems over time were loaded, and the period was specified. Data was then aligned with the metadata, in addition to removing systems with missing metadata and corrupt systems that generated power overnight. Next, the satellite data were loaded. The time series rearranged from date and time to integer sequence of numbers where the distance between two numbers equals to five minutes and mapped with the data. Cloud coverage was measured through average visible HRV level in the area above the PV system. Both cloud coverage and time series were merged with the power generated by the PV system to form the training and testing set for the model.  
\subsection{Model Accuracy Evaluation}
This subsection explores different factors that affect model accuracy. There are several well-known methods for evaluating the performance of a prediction model. Mean square error, root mean square error (RMSE) and MAPE are examples of error-based measurement. In this paper, MAE is used to evaluate the accuracy of the model. It is defined as follows:
\begin{equation}
    MAE(Y,Y*) = \frac{1}{N}\sum_{i=1}^{N} \left|{y_{i}-y*_{i}}\right|
\end{equation}
Mean absolute error measures how much error expected on average from a PV system; in this case, it is measured in watts (W). If the model has a MAE of 100 W trained against a PV system, then the predicted value can differ from the actual value by 100 W on average. The reason why MAE was chosen for this study is that it helps quantify the actual difference in power load generated on average. Four PV systems around the United Kingdom were tested in each trial. The tested PV systems had the numbers 709, 1556, 1627 and 1872, which were randomly selected from a pool of PV systems, and had a capacity of 2,460 W; 3,870 W; 2,820 W and 3,960 W, respectively. The experiments were divided into two sets. The first set of experiments predicted the power generated from PV systems over the next 48 hours by testing different factors, given the cloud coverage at each point.
The first factor to examine from the first set is the training period. The drawback with the Gaussian process is that it does not scale well by increasing the number of observations, in which the training complexity can reach $O(N^3)$. Consequently, selecting the features that will be considered in the model and the number of observations that will be trained are crucial for the training process. In addition, a long training period adds seasonal uncertainty and efficiency reduction over time, which may affect accuracy. Four periods are examined to compare training periods and the accuracy of the model. The periods are one week, two weeks, three weeks and one month. The second factor examined is the average HRV value of the sky area above the PV system. If the sky area considered is too large, then the HRV values will not be relevant and will affect the accuracy of the model. If the coverage area is too small, then the solar irradiance captured by a PV system will also be inaccurate. Each pixel on the map represents 1,000 m on the ground. The tested sky areas were 2 by 2 pixels, 6 by 6 pixels and 12 by 12 pixels. The third factor examined is the kernel. The kernel represents the prior knowledge about the data and how the data are correlated. The kernel has a significant effect on the model's accuracy; therefore, it should be carefully selected.\\
Before discussing different kernel choices, it is essential to note that the white noise kernel is added to the main kernel in all experiments conducted in this paper. This kernel represents the noise in the data by adding uncertainty to the observed data, and it does not change the prediction. The white noise kernel represented by the following:
\begin{equation}
    K_{WN}(x_{i}, x_{j}) = \sigma^{2}\delta(i,j)
\end{equation}
where $\delta$ is the Kronecker delta, and $\sigma^{2}$ is the variance parameter.\\
Due to the sun's movement, power generated from PV systems is periodic in nature. Accordingly, the main kernel in the model is the periodic kernel. However, the latter is a wrap kernel and must have a base kernel that inherits from a stationary kernel. In other words, the periodic kernel is applied in a domain of a given stationary kernel. The periodic kernel with squared exponential stationary kernel is given in (3). The squared exponential (SE) kernel (4), rational quadratic (RQ) kernel (5) and Matern kernel (6) are examined in this paper:
\begin{equation}
\begin{split}
    K_{per-SE}(x_{j},x_{j};h,w,T) = \\
    h^{2} exp  
 \left( -\frac{1}{2w^{2}}\sin^2\left(\pi\left|\frac{x_{i} - x_{j}} {\lambda}\right|\right) \right)
\end{split}
\end{equation}
\begin{equation}
    K_{SE} = h^{2} exp  
 \left[ -\left(\frac{x_{i} - x_{j}} {\lambda}\right)^2  \right]
\end{equation}
\begin{equation}
    K_{RQ}(x_{i}, x_{j}) = h^{2}\left(1+ \frac{(x_{i} - x_{j})^{2}} {\alpha\lambda^{2}}\right)^{-\alpha}
\end{equation}
\begin{equation}
\begin{split}
    K_{M}(x_{i}, x_{j}) = h^{2}\frac{1}{\Gamma(v)2^{v-1}}\left(2\sqrt{v}\frac{|x_{i}-x_{j}|}{\lambda}\right)\\
    \mathbb{B}_{v}\left(2\sqrt{v}\frac{|x_{i}-x_{j}|}{\lambda}\right)
\end{split}
\end{equation}
where $h$ is amplitude, $\lambda$ is the input scale, $T$ is the period, $w$ is the roughness (similar to role of  $\lambda$ in stationary covariances), $\Gamma()$ is s standard Gamma function, $\mathbb{B}()$ is a modified Bessel function of second order, $\alpha$ is known as the index, and $v$ controls the degree of differentiability of the resultant functions \cite{b24}
The first two kernels provide a high degree of freedom with relatively few hyperparameters, whilst the third kernel is better for rougher and less smooth variations.

The second set of experiments predicted the energy generated over the next four hours by fixing the training period to three weeks and the periodic kernel to Matern12 as a base kernel. The experiments compared the accuracy of the predictions against a sky coverage  of 6 by 6 pixels and 12 by 12 pixels with and without cloud coverage input during the specified period. When cloud coverage was not considered, the last observed cloud coverage was used to predict the following four hours.

\section{Results}
Results for the first set of experiments (Table I) demonstrate that MAE was lowest with one week of training and the highest with one month of training. However, one week of training may not be enough to cover different weather situations; therefore, three weeks of training was considered as a period in the next set of tests, as it has the second lowest MAE. Moreover, it is worth mentioning that there was a significant difference in training time between each training period, since many GP tasks have complexity of $O(N^3)$.

\begin{table*}
\caption{Set one Experiments result}
\begin{tabularx}{\textwidth}{@{}l*{10}{C}c@{}}
\toprule
Training Period     & Sky Coverage & Kernel Structure & System 709 &    System 1556   &   System 1627    &   System 1872 & Average (MAE) \\ 
\midrule
1 week  & \multirow{4}{*}{2X2 pixels}      & \multirow{4}{*}{Matern12} & 112.07   & 233.14  & 199.46    & 143.73   & $\boldsymbol{172.1}$\\ 
2 weeks &   &   & 161.9    & 418.8   & 162.1    & 142.15   & 221.24\\ 
3 weeks &   &   & 231.54  & 237.17  & 161.1   & 161.08  & 197.72\\ 
1 month &   &   & 139.38   & 423.79 & 177   & 288.52 & 257.17\\ 
\addlinespace
\multirow{4}{*}{3 weeks}  & 2X2 & \multirow{3}{*}{Matern12} & 231.54   & 237.17  & 161.1    & 161.08   & 197.7225\\ 
        & 6X6 & & 105.94    & 322.23   & 162.5    & 154.79   & 186.365\\ 
    & 12X12 &   & 109.59  & 227.2  & 165.89   & 146.6  &    $\boldsymbol{162.32}$\\ 
\addlinespace
\multirow{3}{*}{3 weeks}  & \multirow{3}{*}{2X2}      & Squared Exponential & 113.705   & 236.67  & 177.54    & 152.13   & 170.01125\\ 
    &   & Rational Quadratic & 124.1    & 223.95   & 709.615    & 147.52   & 301.29625\\ 
    &   & Matern12 & 109.59  & 227.2  & 165.89   & 146.6  & $\boldsymbol{162.32}$\\ 
\bottomrule
\end{tabularx}
\end{table*}

%\begin{figure}[htbp]
%\centerline{\includegraphics[width=0.5\textwidth]{Training Period vs Training Time v2.png}}
%\caption{Comparison between training period and training time}
%\label{Figure 1}
%\end{figure}

\begin{figure}[tbp]
\centerline{\includegraphics[width=0.35\textwidth]{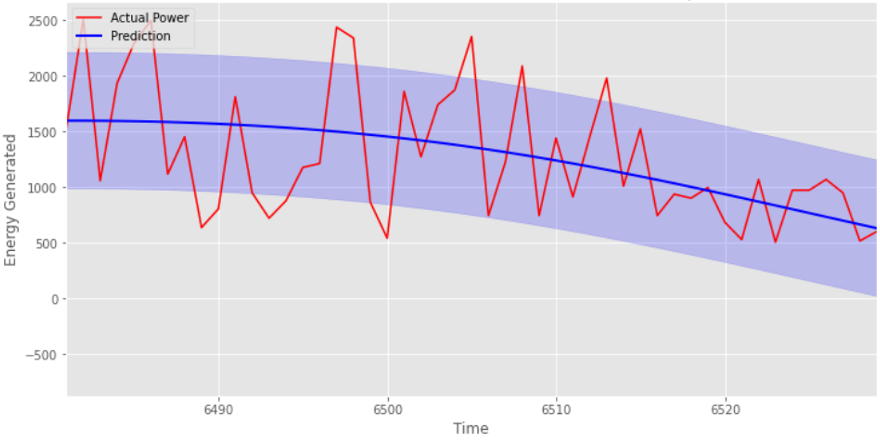}}
\caption{Energy prediction vs.\ actual generation for a PV system with initial cloud input on a cloudy day with scattered clouds, over 4 hours. MAE: 327.41.}
\label{Figure 2}
\end{figure}

On sky coverage, the model performed better with 12 by 12 pixel coverage for a 48-hour forecast, with average MAE of 162.32.
A kernel selection test found the Matern12 kernel to perform best on average. The conducted experiments showed that a three-week training period, a 12 by 12 pixel sky coverage and a periodic kernel with Matern12 base kernel yielded the best results for 48-hour predictions. As the model made use of cloud cover data, the results may be affected by the accuracy of cloud coverage prediction. Figure 3 shows MAE for the recommended settings per tested days.

\begin{figure}[tbp]
\centerline{\includegraphics[width=0.35\textwidth]{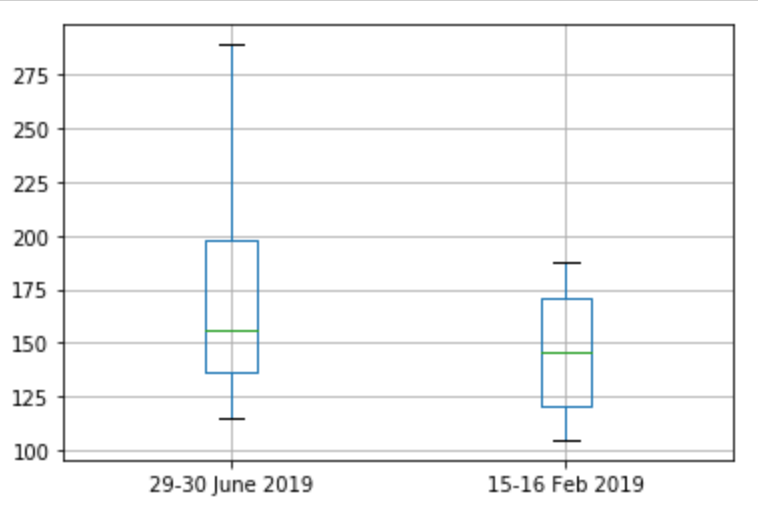}}
\caption{MAE box plot for 48-hours forecasts presented by testing periods}
\label{Figure 3}
\end{figure}

\begin{table}[tbp]
\caption{Set two Experiments result}
\begin{tabular}{c|cc|cc}
\multirow{2}{*}{\textbf{System Number}} & \multicolumn{2}{c|}{\textbf{With Cloud Coverage}} & \multicolumn{2}{c|}{\textbf{Without Cloud Coverage}} \\
\cline{2-5}
    & \textbf{6X6} & \textbf{12X12} & \textbf{6X6} & \textbf{12X12}\\
\hline
709 & 145.45 & 285.99 & 322.98 & 327.67\\
1556 & 457.81 & 842.47 & 527.75 & 837.665\\
1627 & 180.74 & 192.12 & 196.57 & 202.56\\
1872 & 240.52 & 224.62 & 317.97 & 320.91\\
\hline
\textbf{Average} & \textbf{256.04} & \textbf{386.3} & \textbf{341.31} & \textbf{422.2}\\
\end{tabular}
\end{table}

We also tested the difference in accuracy with and without cloud cover information, with a much shorter forecasting period of four hours (Table II, Figure 3). We analysed four selected days, each day featuring different weather conditions. The lowest MAE was in a clear sky and highest with scattered clouds. The model was able to follow the general trend in all cases, but cloud cover improved predictions. Note that for PV system 1556, errors were generally larger than for others.

\begin{figure}[t]
\centerline{\includegraphics[width=0.4\textwidth,clip,trim=0mm 8mm 0mm 0mm]{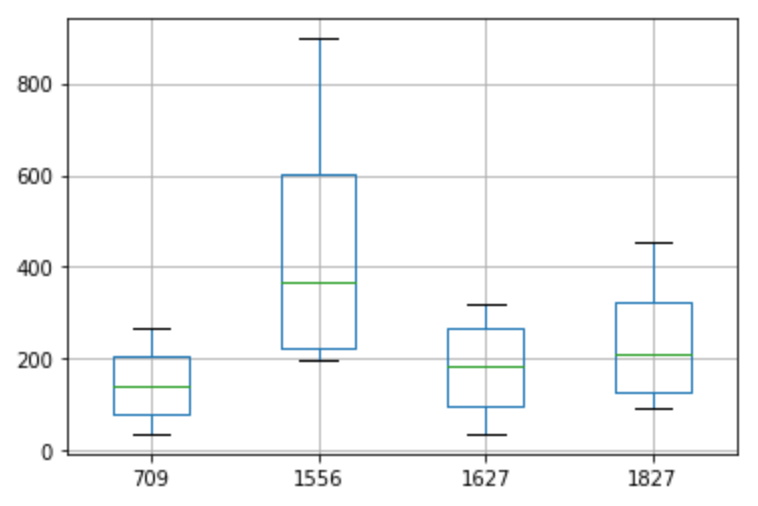}}
\centerline{\includegraphics[width=0.4\textwidth]{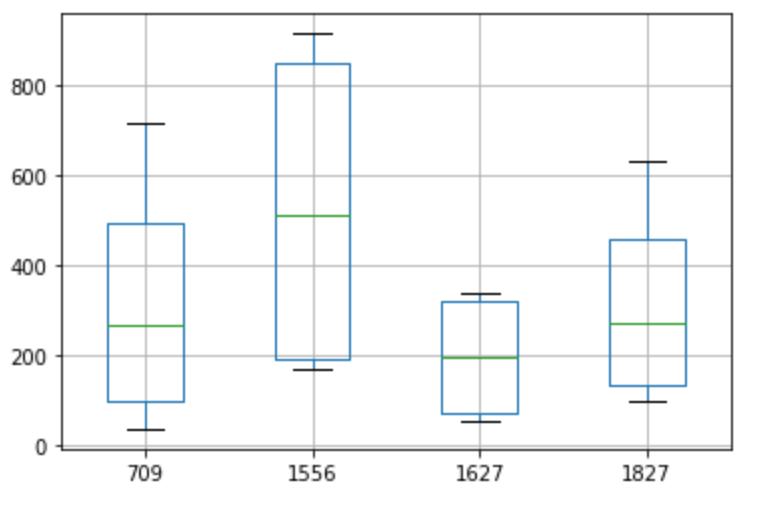}}
\caption{MAE box plot with outliers for four-hour forecasts. Upper: with consideration of cloud coverage presented to the PV system. Lower: without.}
\label{Figure 4}
\label{Figure 5}
\end{figure}

\section{Conclusion}
This paper evaluated three different factors which affect the GPR model when predicting power generated by PV systems: the training period, sky coverage area and kernel. 
The main advantage of a GP model is that not only can it predict the amount of power generated, but it also captures the uncertainty of the predicted value at each point. 
The analysis shows that the Matern12 kernel is the best one amongst the tested kernels, as it was more flexible and could  capture the uncertainty presented in the generated power. Furthermore, a training period of three weeks and a sky coverage area of 12 by 12 pixels were the best choices for 48-hour forecasts. In comparison, a sky coverage area of 6 by 6 pixels was preferable for four-hour forecasts.

The results show that the model was able to accurately predict power generated by PV systems when cloud coverage was stable. However, the model suffered from tracking very rapid changes in cloud coverage when the clouds were scattered. The results also confirmed that the main factor for short-term predictions of solar power is solar irradiance and that cloud coverage has a significant effect on power production. %Comparing the results discussed in this paper with other papers might not be accurate as system performance and weather conditions are different.

\vspace{12pt}

\clearpage
\appendix
\section{Gaussian process}
Gaussian process (GP) is a probability distribution over possible functions that fit a set of points. Having a joint distribution of these random variables generates a multivariate Gaussian distribution. A univariate Gaussian distribution can be described by mean and variance.  

\begin{equation}
    \mathcal{N}\sim(\mu,\sigma^2)
\end{equation}
where $\mathcal{N}$ represents Gaussian distribution with mean $\mu$ and variance $\sigma^2$.\\
The joint univariate Gaussian distribution is described instead by mean and a covariance matrix.
\begin{equation}
    \mathcal{N}\sim(\mu,\Sigma)
\end{equation}
where $\mathcal{N}$ represents Gaussian distribution with mean $\mu$ and covariance $\Sigma$.\\
The covariance matrix will contains the variance of each random variable, which expressed diagonally in the matrix, and the covariance between the random variables. The covariance between variables represents how two variables are correlated.
Assuming that we have two-variable Gaussian distribution with variables $y_{1}, y_{2}$ having $\mu = 0$ and covariance matrix 
$\begin{bmatrix}
1 & 0.9\\
0.9 & 1
\end{bmatrix}$.
Figure 6 shows the possible values that $y_{2}$ can have given $y_{1}$, described by the red Gaussian distribution as $y_{1}$ and $y_{2}$ are correlated.
\begin{figure}[htbp]
\centerline{\includegraphics[width=0.5\textwidth]{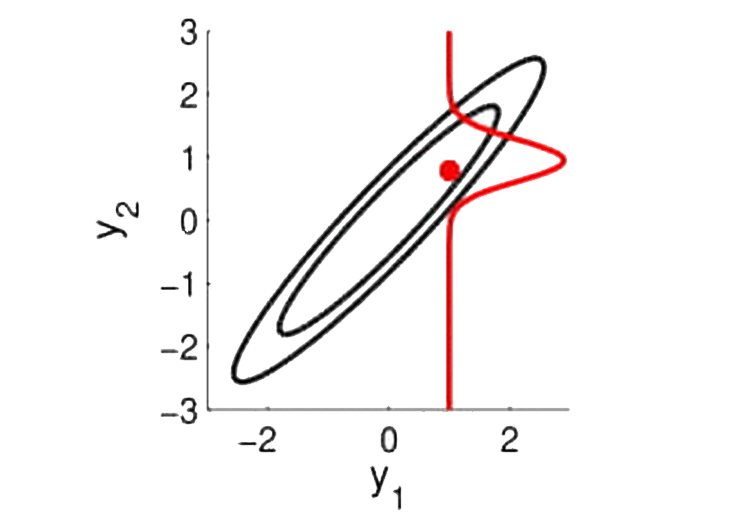}}
\caption{Possible values for $y_{2}$ given $y_{1}$}
\label{Figure 6}
\end{figure}

Figure 7 shows an example of a ten-dimensional posterior distribution, with observations on locations 2, 6, and 8. The black line represents the mean and the grey area represents the standard deviation.
\begin{figure}[htbp]
\centerline{\includegraphics[width=0.5\textwidth]{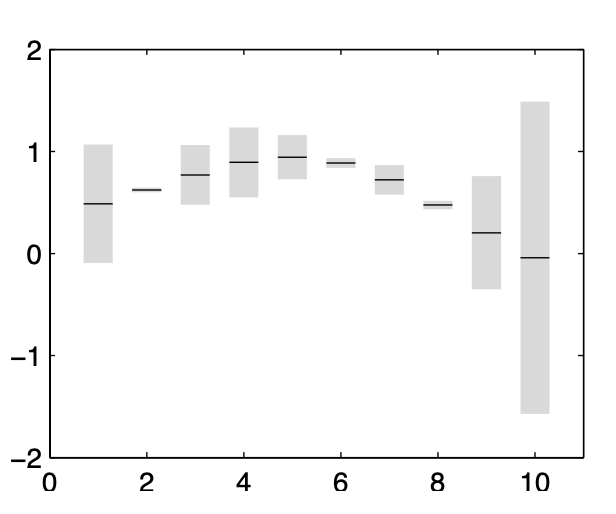}}
\caption{10 dimensional posterior distribution with observations on locations 2, 6, and 8 \cite{b24}}
\label{Figure 7}
\end{figure}
\\Thus far, what has been presented is a discrete set of predictions and observations points. However, in order to model observations with real values, we require arbitrarily large sets and must calculate its covariance or otherwise have the covariance generated by the covariance kernel function, which will provide the covariance element between any arbitrary samples. Choosing the covariance kernel function insufficient on its own. Other properties, such as the right length scale that represents the data more accurately, can still be vague. Such properties are called hyperparameters, and they define the shape of the covariance kernel function. Finding the best values for these hyperparameters that represent the given data is called 'training the model'.\\
For a set of random variables, such as  $X = \{x_{1}, x_{2}, ..., x_{n}\}$, then we can define the covariance matrix can be defined as 
\begin{equation}
\boldsymbol{K}(X,X) = 
\begin{pmatrix}
k(x_{1},x_{1}) & k(x_{1},x_{2}) & \dotsm & k(x_{1},x_{n})\\
k(x_{2},x_{1}) & k(x_{2},x_{2}) & \dotsm & k(x_{2},x_{n})\\
\vdots & \vdots & \vdots & \vdots\\
k(x_{n},x_{1}) & k(x_{n},x_{2}) & \dotsm & k(x_{n},x_{n})
\end{pmatrix}
\end{equation}
The entire function evaluation associated with points in $X$ is a draw from Multivariate Gaussian distribution.
\begin{equation}
    p(Y(X)) = \mathcal{N} (\mu(X),K(X,X))
\end{equation}
where $Y = \{y_{1}, y_{2}, \dotsm, y_{n}\}$ are dependent function values, evaluated with random variables $X = \{x_{1}, x_{2}, ..., x_{n}\}$.\\
Using the Multivariate Gaussian distribution we can obtain posterior distribution over $\boldsymbol{y*}$ for test datum $\boldsymbol{x*}$ given by 
\begin{equation}
    p(\boldsymbol{y*}) = \mathcal{N}(\boldsymbol{m*},\boldsymbol{C*})
\end{equation}
where,
\begin{equation}
    \boldsymbol{m*} = \boldsymbol{\mu}(X*) + \boldsymbol{K}(X*,X)\boldsymbol{K}(X,X)^{-1} (Y(X)-\mu(X)),
\end{equation}
\begin{equation}
    \boldsymbol{C*} = \boldsymbol{K}(X*,X*) - \boldsymbol{K}(X*,X)\boldsymbol{K}(X,X)^{-1} \boldsymbol{K}(X*,X)^T.
\end{equation}
Figure 8 is an example of Gaussian process plot in which the red points represent  observations and the blue line represents the function that best fits the observations. The blue area represents the uncertainty of the output \cite{b6}\cite{b21}\cite{b22}\cite{b24}.\\
\begin{figure}[htbp]
\centerline{\includegraphics[width=0.5\textwidth]{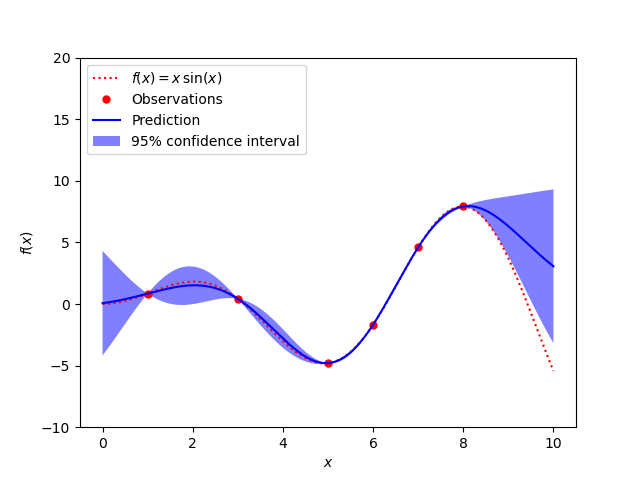}}
\caption{Example of gaussian Process output \cite{b23}}
\label{Figure 8}
\end{figure}

\end{document}